\documentclass[journal,twoside,web]{IEEEtran}
\usepackage{cite}
\usepackage{dutchcal}
\usepackage{amsmath,amssymb,amsfonts}
\usepackage{mathrsfs}
\usepackage{algorithmic}
\usepackage[linesnumbered,ruled]{algorithm2e}
\usepackage{hyperref}       
\usepackage{url}            
\usepackage{booktabs}       
\usepackage{amsfonts}       
\usepackage{nicefrac} 
\usepackage{microtype}      
\usepackage{lipsum}
\usepackage{graphicx}
\usepackage{textcomp}
\usepackage{float}
\usepackage{makecell}
\usepackage{multirow}
\usepackage{color}
\usepackage{diagbox}
\usepackage{color,soul}
\usepackage{gensymb}
\def\etal{\emph{et al.}}
\def\ie{\emph{i.e.}}

\begin{document}
\title{RCPS: Rectified Contrastive Pseudo Supervision for Semi-Supervised Medical Image Segmentation}
\author{Xiangyu Zhao, Zengxin Qi, Sheng Wang, Qian Wang, Xuehai Wu, Ying Mao, and Lichi Zhang
\thanks{This work is partially supported by National Natural Science Foundation of China (NSFC) grants (Grant No. 62001292), Shanghai Municipal Science and Technology Major Project (No.2018SHZDZX01), ZJ Lab, and Shanghai Center for Brain Science and Brain-Inspired Technology, Shanghai Zhou Liangfu Medical Development Foundation “Brain Science and Brain Diseases Youth Innovation Program”, the Lingang Laboratory, Grant No. LG202105-02-03.}
\thanks{Corresponding authors: Y. Mao, and L. Zhang (e-mails: maoying@fudan.edu.cn, lichizhang@sjtu.edu.cn).}
\thanks{X. Zhao and Z. Qi contributed equally to this work.}
\thanks{X. Zhao, S. Wang and L. Zhang are with School of Biomedical Engineering, Shanghai Jiao Tong University, Shanghai, 200030, China (e-mails: \{xiangyu.zhao, wsheng, lichizhang\}@sjtu.edu.cn).}
\thanks{X. Wu, Z. Qi, and Y. Mao are with Department of Neurosurgery, Huashan Hospital, Shanghai Medical College, Fudan University, also with National Center for Neurological Disorders, also with Shanghai Key Laboratory of Brain Function and Restoration and Neural Regeneration, and also with State Key Laboratory of Medical Neurobiology and MOE Frontiers Center for Brain Science, School of Basic Medical Sciences and Institutes of Brain Science, Fudan University (e-mails: wuxuehai2013@163.com, \{qizengxin, maoying\}@huashan.org.cn)}
\thanks{Q. Wang is with School of Biomedical Engineering, ShanghaiTech University, Shanghai, 201210, China, 
and with Shanghai Clinical Research and Trial Center, Shanghai, 201210, China (e-mail: qianwang@shanghaitech.edu.cn).}
}
\maketitle
\bibliographystyle{ieeetr}

\begin{abstract}
Medical image segmentation methods are generally designed as fully-supervised to guarantee model performance, which requires a significant amount of expert annotated samples that are high-cost and laborious. Semi-supervised image segmentation can alleviate the problem by utilizing a large number of unlabeled images along with limited labeled images. 
However, learning a robust representation from numerous unlabeled images remains challenging due to potential noise in pseudo labels and insufficient class separability in feature space, which undermines the performance of current semi-supervised segmentation approaches.
To address the issues above, we propose a novel semi-supervised segmentation method named as Rectified Contrastive Pseudo Supervision (RCPS), which combines a rectified pseudo supervision and voxel-level contrastive learning to improve the effectiveness of semi-supervised segmentation. 
Particularly, we design a novel rectification strategy for the pseudo supervision method based on uncertainty estimation and consistency regularization to reduce the noise influence in pseudo labels.
Furthermore, we introduce a bidirectional voxel contrastive loss in the network to ensure intra-class consistency and inter-class contrast in feature space, which increases class separability in the segmentation. The proposed RCPS segmentation method has been validated on two public datasets and an in-house clinical dataset. Experimental results reveal that the proposed method yields better segmentation performance compared with the state-of-the-art methods in semi-supervised medical image segmentation. The source code is available at \url{https://github.com/hsiangyuzhao/RCPS}.
\end{abstract}

\begin{IEEEkeywords}
Medical image segmentation, semi-supervised learning, contrastive learning, pseudo supervision
\end{IEEEkeywords}

\section{Introduction}
\label{sec:introduction}
Accurate and automatic segmentation of anatomical structures or lesions in medical images is highly desirable in various clinical applications, including image-guided intervention, radiation therapy, and computer-aided diagnosis \cite{kaur2017survey}. The rapid developments of deep learning have led to the emergence of numerous image segmentation methods for the quantitative analysis of medical images. However, these methods often follow a fully-supervised manner, which requires a large amount of well-annotated data to achieve satisfactory performance. Meanwhile, manual annotations of medical images are high-cost and laborious, particularly for 3D images such as CT and MRI scans. Therefore, it is crucial to develop automatic segmentation methods that can reduce the demand for extensive training datasets.

Semi-supervised segmentation has emerged as a promising technique to tackle the aforementioned issues, as it enables the utilization of unlabeled data to enhance segmentation performance with only a limited number of labeled samples. Since the ground truths of unlabeled samples are unavailable, pseudo supervision has emerged as a popular strategy for semi-supervised learning, where the predictions of unlabeled images by the segmentation network are used as pseudo labels to supervise the training \cite{yang2022survey}. Semi-supervised learning based on pseudo supervision is categorized into two distinct types: offline and online training. 
Offline training involves simple pseudo labeling \cite{lee2013pseudo} of the unlabeled data to expand the training dataset, while online training introduces perturbations from the models \cite{ouali2020semi, chen2021semi} or the input data \cite{tarvainen2017mean, sohn2020fixmatch, zou2020pseudoseg, zhong2021pixel} based on consistency learning, and encourages the similarity of the predictions between the perturbed inputs and the pseudo labels.

Despite the accomplishments in the prior studies, two major challenges still need to be addressed in the task of semi-supervised medical image segmentation: 
(1) \emph{Noisy nature of pseudo labels in semi-supervised learning}: As previously introduced, pseudo supervision such as \cite{zou2020pseudoseg, zhong2021pixel} is implemented by utilizing the segmentation model to produce pseudo labels for the unlabeled images. However, the segmentation model is usually susceptible to label noise, which can inevitably hinder the effectiveness of pseudo supervision.
(2) \emph{Insufficient supervision in the feature space}: Current semi-supervised learning methods such as \cite{ouali2020semi, chen2021semi, tarvainen2017mean, sohn2020fixmatch, zou2020pseudoseg} only provide supervision in the label space, but they lack explicit supervision in the feature space to further improve class separability.

To tackle these challenges, here we propose a Rectified Contrastive Pseudo Supervision (RCPS) method for semi-supervised segmentation in medical images. The major contribution of this work is two-fold: (1) We develop a novel rectified pseudo supervision strategy which generates two augmented views with different appearances from the original input, and introduces the rectification of pseudo labels based on uncertainty estimation and consistency regularization in model predictions. In this way, we force the model to learn robustly under the label noise in pseudo supervision. In addition, we design a novel triplet learning strategy to implement both uncertainty rectification and consistency regularization simultaneously, which can significantly enhance the capacity of our semi-supervised segmentation compared with previous attempts in \cite{luo2021efficient, wu2022mutual}. (2) We introduce a bidirectional voxel contrastive learning strategy to address the issue of insufficient class separability. It can further optimize the model to learn better class separability by pulling the voxels belonging to the same class together in the feature space, while pushing those in different classes apart. Furthermore, our proposed bidirectional voxel contrastive learning offers a much stronger supervision signal by bidirectional computation than the previous attempts in \cite{zhong2021pixel, you2022simcvd}.

We validate the proposed RCPS on two public benchmark datasets: left atrial cavity segmentation in MRI \cite{xiong2021global} and pancreas segmentation in CT \cite{nihct}. Experimental results demonstrate the superiority of the proposed RCPS to other state-of-the-art methods for both benchmarks. We also evaluate RCPS on an in-house clinical dataset of brain segmentation for patients with traumatic brain injuries (TBI), which achieves state-of-the-art segmentation performance and surpasses the previous works.

To sum up, the contributions of this work consist of the following aspects:
\begin{itemize}
    \item We design a novel semi-supervised segmentation method that generates two different augmented views from the original input and applies a triplet learning strategy, aiming at addressing the noisy nature of pseudo supervision and insufficient feature space supervision in semi-supervised medical image segmentation.
    \item We introduce rectified pseudo supervision to improve robustness under variant appearances in the image space. Our approach employs prediction uncertainty estimation and consistency regularization to alleviate the noise in pseudo supervision and achieve better efficiency than previous attempts.
    \item We propose a bidirectional voxel contrastive loss with a novel confident negative sampling strategy. This loss encourages the two augmented views to be similar and both views to be far away from the negative samples in the feature space. We extend contrastive learning to both voxel space and bidirectional computation, which effectively enlarges the separability between different semantic classes in feature space.
    \item We validate the proposed RCPS on two public benchmark datasets and an in-house clinical dataset. Experimental results show that RCPS outperforms the state-of-the-art methods, which exhibits potential for implementation in clinical scenarios.
\end{itemize}

\section{Related Works}
\subsection{Medical Image Segmentation}
The development of deep learning has significantly improved the accuracy of semantic segmentation \cite{ronneberger2015u, milletari2016v, long2015fully}. In the field of medical image segmentation, U-Net \cite{ronneberger2015u} and its variants \cite{milletari2016v, zhou2018unet++, oktay2018attention} are widely adopted due to their efficiency and accuracy. The main strength of U-Net lies in its design of skip connections, and its variants continue to enhance performance by exploring better feature representations through modification of network topology \cite{milletari2016v, zhou2018unet++} or feature map refinement using attention mechanism \cite{oktay2018attention, zhao2022prior}. Recently, transformer networks \cite{vaswani2017attention} have received much attention in medical image segmentation. Chen \etal \cite{chen2021transunet} integrated a transformer into the U-Net encoder to enhance the model's capacity, thus improving segmentation performance. Cao \etal \cite{cao2021swin} developed a pure transformer U-Net-like network named Swin-Unet, which replaces convolutional blocks with transformer blocks for better feature extraction. Despite the success of both CNNs and transformers in medical image segmentation, their design is generally based on fully-supervised learning. Therefore, the performance improvements of fully-supervised segmentation are restricted due to the limited number of labeled training samples.
\subsection{Semi-Supervised Learning}
\label{sec:related_ssl}
Semi-supervised learning algorithms are based on three core assumptions \cite{van2020survey}: (1) \textit{smoothness assumption}: similar inputs should yield similar outputs and vice versa; (2) \textit{low-density assumption}: samples of the same class tend to form a cluster in the feature space, and decision boundaries of different classes should only pass through the low-density areas in the feature space; (3) \textit{manifold assumption}: samples located on the same low-dimensional manifold should belong to the same class. Most semi-supervised segmentation approaches based on pseudo supervision follow these assumptions and can be divided into two categories, as discussed in Section~\ref{sec:introduction}. The first category utilizes direct pseudo labeling to expand the training dataset. Lee \etal \cite{lee2013pseudo} proposed to use the predictions of the fully-supervised algorithm as the pseudo labels of the unlabeled data. However, simple pseudo-labeling can introduce much label noise to the training process. To solve this, Sohn \etal \cite{sohn2020fixmatch} proposed to threshold the model predictions and only preserve those with high confidence, thereby reducing the false label ratio. Additionally, Zheng \etal \cite{zheng2021rectifying} estimated the uncertainty of the model predictions by calculating the KL-divergence between the main output and the auxiliary output, enabling adaptive thresholding of pseudo labels. The second category utilizes the consistency of pseudo labels under different perturbations. For instance, Tarvainen \etal \cite{tarvainen2017mean} proposed to generate two different augmented views of the same input and use the output of the teacher model as supervision for the student model. Ouali \etal \cite{ouali2020semi} proposed a segmentation model with multiple auxiliary decoders and encouraged consistency between the predictions made by the main decoder and the auxiliary decoders. Chen \etal \cite{chen2021semi} proposed cross-pseudo supervision by enforcing pseudo supervision between the predictions of two models with different parameter initialization.

The above techniques are also popular in medical image segmentation. The variants of the mean teacher framework have been widely applied in medical image segmentation \cite{cui2019semi, yu2019uncertainty, cao2020uncertainty}. Co-training \cite{luo2021teach} and multi-task learning \cite{li2020shape, luo2021dualtask, you2022simcvd} are also frequently employed to explore the consistency of different models or tasks. Furthermore, uncertainty rectification \cite{luo2021efficient, wu2022mutual, luo2022semi} is generally used to improve model confidence. 
\subsection{Contrastive Learning}
Contrastive learning is widely adopted in self-supervised learning, which explores learning discriminative feature representations without any annotations \cite{krishnan2022self}. The key idea of contrastive learning is to learn distinctive feature representations to enlarge the margin between different classes in the feature space, allowing for the discrimination of positive and negative pairs. To achieve this, the number of negative samples should be large enough to enable the model to learn the distinction between different data pairs. Existing methods typically maintain a large mini-batch \cite{chen2020simple, chen2020big} during training or use a memory bank \cite{he2020momentum, chen2020improved, chen2021empirical} updated by momentum to hold a significant number of negative samples. The success of contrastive learning in image-level tasks has motivated researchers to transfer it to dense prediction tasks. For instance, Wang \etal \cite{wang2021exploring} modified image-level contrastive learning to the voxel level to improve segmentation performance. Pixel contrastive learning has also been broadly adopted for semi-supervised segmentation \cite{zhong2021pixel, lai2021semi, you2022simcvd} beyond full supervision. Zhong \etal \cite{zhong2021pixel} introduced pixel contrastive learning with confidence sampling to consistency training in semi-supervised segmentation for improving the segmentation capacity. Lai \etal \cite{lai2021semi} proposed to generate two different views from the same input and encourage consistency in the overlapped regions while establishing bidirectional contrast in other regions. 
\begin{figure*}[ht]
	\centering
	\includegraphics[width=\textwidth]{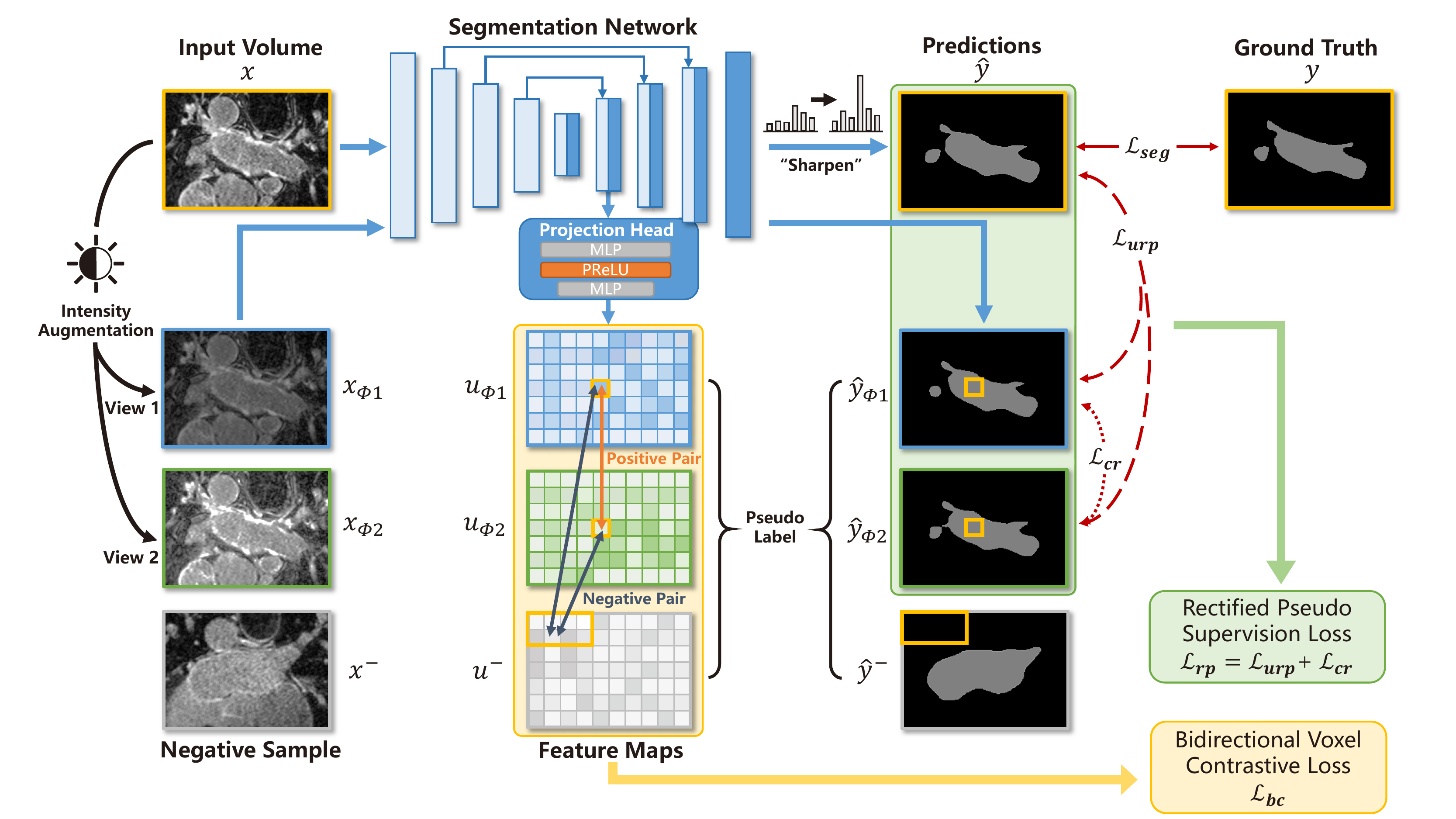} 
	\caption{The workflow of the proposed RCPS. Given an input $x$, the proposed RCPS performs two augmentations on the original input and generates two augmented views $x_{\phi 1}$ and $x_{\phi 2}$. Rectified pseudo supervised loss and bidirectional voxel contrastive loss are calculated on both labeled and unlabeled data. The scheme is drawn in 2D in favor of comprehension.}
	\label{fig:scheme}
\end{figure*}
\section{Methods}
\label{sec:methods}
In this section, we introduce RCPS, a semi-supervised learning method by leveraging limited labeled data and ample unlabeled data for improving segmentation performance in medical images. We first present an overview of the proposed RCPS method, and then describe the learning strategies utilized in RCPS, which include rectified pseudo supervision and bidirectional voxel contrastive learning.
\subsection{Overview of the Framework}
\label{sec:overview}
In semi-supervised segmentation, we assume a training dataset with a small labeled subset containing $N$ labeled data and a large unlabeled set containing $M$ unlabeled data, where $M \gg N$. For convenience, we denote the labeled subset as $D_l=\{(x_i^l, y_i^l)\}_{i=1}^{N}$ and the unlabeled set as $D_u=\{(x_i^u)\}_{i=1}^{M}$, where $x_i \in \mathbb{R}^{H \times W \times D}$ denotes the training image, and $y_i \in \mathbb{B}^{C \times H \times W \times D}$ denotes the training label (if available). The training objective of semi-supervised segmentation is to leverage the images in unlabeled subset $D_u$ to improve the segmentation performance, compared with training with the labeled subset $D_l$ only.

The proposed RCPS is based on the U-Net architecture \cite{ronneberger2015u}, with a projection head in the second upsampling block, which offers a delicate balance between semantic richness and computational cost for calculating contrastive loss, as shown in Fig. \ref{fig:scheme}.
The framework consists of four data inputs in one forward pass: one original data input $x$, two augmented views $x_{\phi 1}$ and $x_{\phi 2}$ transformed from $x$, and one negative input $x^-$ for bidirectional voxel contrastive learning. The softmax outputs of these inputs are denoted as $\hat{y}$, $\hat{y}_{\phi 1}$, $\hat{y}_{\phi 2}$ and $\hat{y}^-$, respectively. Also, for bidirectional voxel contrastive learning, the intermediate features of $x_{\phi 1}$, $x_{\phi 2}$ and $x^-$ are outputed by the projection head as well, denoted as $u_{\phi 1}$, $u_{\phi 2}$ and $u^-$, respectively.

For the labeled data, we use the combination of cross entropy loss and Dice loss \cite{milletari2016v} to supervise the training, which is popular in medical image segmentation:
\begin{equation}
\label{eqn1}
    \mathcal{L}_{seg}(\hat{y}, y) = \mathcal{L}_{ce}(\hat{y}, y)+\mathcal{L}_{Dice}(\hat{y}, y),
\end{equation}
where $\mathcal{L}_{ce}$ denotes the cross entropy loss, $\mathcal{L}_{Dice}$ denotes the Dice loss.

For unlabeled data, we calculate the rectified pseudo supervision loss $\mathcal{L}_{rp}$ and bidirectional voxel contrastive loss $\mathcal{L}_{bc}$, which are discussed in Sec. \ref{sec:rectified} and Sec. \ref{sec:contrastive}, respectively. The overall unsupervised loss $\ell_{unsup}$ is calculated as below:
\begin{equation}
\label{eqn2}
\ell_{unsup}=\alpha \mathcal{L}_{rp}(\hat{y}_{\phi 1}, \hat{y}_{\phi 2}, \hat{y})+\beta \mathcal{L}_{bc}(u_{\phi 1}, u_{\phi 2}, u^-)
\end{equation}
where $\alpha$ and $\beta$ are hyperparameters to balance the loss, depending on the specific task.

Note that, unlike previous works which only add the extra unsupervised losses \cite{luo2022semi, wu2022mutual, you2022simcvd}, the proposed rectified pseudo supervision loss and the bidirectional voxel contrastive loss are added to the labeled data as well. Thus, the overall supervised loss $\ell_{sup}$ is calculated as below:
\begin{equation}
\label{eqn3}
\ell_{sup}=\mathcal{L}_{seg}(\hat{y}, y) + \alpha \mathcal{L}_{rp}(\hat{y}_{\phi 1}, \hat{y}_{\phi 2}, \hat{y})+\beta \mathcal{L}_{bc}(u_{\phi 1}, u_{\phi 2}, u^-),
\end{equation}
where $\alpha$ and $\beta$ are identical to the unsupervised settings.
\subsection{Rectified Pseudo Supervision}
\label{sec:rectified}
Since medical images vary in appearance, we adopt pseudo supervision strategy in line with previous works \cite{lee2013pseudo, sohn2020fixmatch, zou2020pseudoseg} in semi-supervised learning. Specifically, we produce two augmented views from the original input by intensity transform including brightness shift, contrast shift and random noising, which can simulate the variance of medical images. 
Pseudo supervision is operated between the original input and augmented views of the original input to learn against the perturbations on image intensity. However, pseudo supervision can be noisy due to the prediction error in pseudo labels, which would hurt model performance subsequently. We propose to rectify pseudo supervision by prediction uncertainty estimation and consistency regularization. The illustration has been made in Fig. \ref{fig:scheme}
\subsubsection{Pseudo Supervision}
Given a data input $x$, we perform two different intensity augmentations and acquire two different augmented views $x_{\phi 1}$ and $x_{\phi 2}$. The augmentation used involves random intensity scaling, random intensity shifting, and Gaussian noising.

Pseudo supervision is performed by applying supervised loss functions between $\hat{y}$ and $\hat{y}_{\phi i} \ (i=1,2)$. By doing so, we encourage the model to learn a robust representation despite the difference in data appearance. In practice, we sharpen the softmax probabilities $\hat{y}$ by dividing the logits (model outputs before softmax activation) by a temperature hyperparameter $T$, to ensure the low-density assumption in semi-supervised learning, which has been discussed in Sec. \ref{sec:related_ssl}. Such an operation could make the pseudo label "harder" and avoid class overlap. Thus, pseudo supervision loss is defined as follows:
\begin{equation}
\label{eqn4}
    \mathcal{L}_{p}(\hat{y}_{\phi i}, \hat{y}) = \mathcal{L}_{ce}(\hat{y}_{\phi i}, \sigma(z / T)),
\end{equation}
where $z$ denotes the logits of $x$, $\sigma$ denotes the softmax function, and $T$ is the temperature. 
\subsubsection{Uncertainty Estimation}
\label{sec:uncertainty}
Direct pseudo supervision between model predictions can be unreliable sometimes, as errors can occur in model predictions, which accounts for the noisy nature of pseudo supervision. Prediction uncertainty estimation and consistency regularization have been introduced to pseudo supervision, to alleviate the effects of the label noise in pseudo supervision.

Pseudo supervision is vulnerable to label noise, thus thresholding the pseudo label with high confidence is a common method \cite{sohn2020fixmatch} to reduce label noise. However, simply thresholding the probabilities by a hard threshold is not suitable for the segmentation task, as the segmentation difficulty of different semantic classes is variable. Hard thresholding can make the difficult classes even harder to pass the threshold and thus lead to a biased prediction that deviates to the background class. 

We propose to use the uncertainty of model predictions to rectify the pseudo supervision, which is computed by KL-divergence \cite{zheng2021rectifying}. With uncertainty estimation, the uncertainty rectified pseudo supervision loss is defined as follows:
\begin{equation}
\label{eqn5}
    \mathcal{L}_{urp}(\hat{y}_{\phi i}, \hat{y})=e^{{-\mathcal{D}_{kl}(\hat{y}_{\phi i}, \hat{y})}}\mathcal{L}_{p}(\hat{y}_{\phi i}, \hat{y})+\mathcal{D}_{kl}(\hat{y}_{\phi i}, \hat{y}),
\end{equation}
\begin{equation}
\label{eqn6}
    \mathcal{D}_{kl}(\hat{y}_{\phi i}, \hat{y})=\hat{y}\log(\frac{\hat{y}}{\hat{y}_{\phi i}}).
\end{equation}

Uncertainty estimation introduces an adaptive voxel-wise weighting to the pseudo supervision loss, where the confident voxels have higher weight and less confident voxels have lower weights. Such rectification reduces the effect of label noise and improves the robustness of segmentation.
\subsubsection{Consistency Regularization}
\label{sec:consistency}
Based on the smoothness assumption in semi-supervised learning, we expect the predictions of $x_{\phi 1}$ and $x_{\phi 2}$ should be similar, despite the different appearances generated by intensity transform. Thus, we introduce consistency regularization to further rectify the pseudo supervision, in order to minimize the disagreement between $\hat{y}_{\phi 1}$ and $\hat{y}_{\phi 2}$. We adopt the cosine distance between the predictions from the two different views:
\begin{equation}
\label{eqn7}
    \mathcal{L}_{cr}(\hat{y}_{\phi 1}, \hat{y}_{\phi 2})=1-\cos(\hat{y}_{\phi 1}, \hat{y}_{\phi 2}),
\end{equation}
\begin{equation}
\label{eqn8}
    \cos(\hat{y}_{\phi 1}, \hat{y}_{\phi 2})=\frac{{\hat{y}_{\phi 1}} \cdot \hat{y}_{\phi 2}}{\parallel \hat{y}_{\phi 1}\parallel_2 \cdot \parallel \hat{y}_{\phi 2}\parallel_2}.
\end{equation}

Minimizing pseudo supervision loss can implicitly reduce the cosine distance between the predictions from two views, but an explicit loss term contributes more regularization during training, which helps stabilize the training process.

The final pseudo supervision loss is the linear combination of uncertainty rectified pseudo supervision loss and consistency regularization.
\begin{equation}
\label{eqn9}
\begin{aligned}
    \mathcal{L}_{rp}(\hat{y}_{\phi 1}, \hat{y}_{\phi 2}, \hat{y})&=\mathcal{L}_{urp}(\hat{y}_{\phi 1}, \hat{y}) + \mathcal{L}_{urp}(\hat{y}_{\phi 2}, \hat{y}) \\
    & + \mathcal{L}_{cr}(\hat{y}_{\phi 1}, \hat{y}_{\phi 2}).
\end{aligned}
\end{equation}

In practice, we detach $\hat{y}$ from the computational graph to stop the gradient in this branch so that the gradients only backpropagate to the branches of augmented views.
\subsection{Bidirectional Voxel Contrastive Learning}
\label{sec:contrastive}
Based on the smoothness assumption in semi-supervised learning, we expect feature space consistency apart from label space consistency. This could be achieved by contrastive learning in the voxel level.
Inspired by existing works, we develop a bidirectional voxel contrastive loss based on InfoNCE loss \cite{he2020momentum} to accomplish this, which is shown in Fig. \ref{fig:scheme}. Voxel pairs at the spatially corresponding locations from $u_{\phi 1}$ and $u_{\phi 2}$ are regarded as positives, as they are augmented from the same original input and should be similar despite the perturbations from intensity transform. The positives are pulled together to decrease intra-class distance and pushed away from the negative ones to ensure a large margin between different classes. 
Formally, given an anchor voxel $\psi_{1} \in u_{\phi1}$, the voxel contrastive loss is calculated as follows:
\begin{equation}
\label{eqn10}
    \mathcal{L}_{c}(\psi_1, \psi_2)=-\log \frac{e^{\cos(\psi_1, \psi_2) / \tau}}{e^{\cos(\psi_1, \psi_2) / \tau} + \sum\limits_{\psi_n \in u^-, n=1}^{N} e^{\cos(\psi_1, \psi_n) / \tau}},
\end{equation}
where $\psi_2$ denotes the positive voxel from the feature map $u_{\phi 2}$ extracted from $x_{\phi 2}$, $\psi_n$ denote the negative voxels, $u^-$ denotes the feature map that contains negative samples, $N$ is the number of sampled negative voxels, and $\tau$ is the temperature hyperparameter.

It should be noted that both augmented views should be pushed away from negative samples. Thus, we calculate another contrastive loss by exchanging the positions of $\psi_1$ and $\psi_2$, which accounts for "bidirectional" calculation. Hence, the bidirectional voxel contrastive loss is the linear summation of two voxel contrastive losses:
\begin{equation}
\label{eqn11}
    \mathcal{L}_{bc}(\psi_1, \psi_2)=\mathcal{L}_{c}(\psi_1, \psi_2)+\mathcal{L}_{c}(\psi_2, \psi_1).
\end{equation}

The key problem for voxel level contrastive learning in image segmentation is the sampling strategy of negative samples. In image segmentation, different voxels in the same image can belong to different classes, especially for background classes which usually cover a relatively large area. Thus, random sampling may sample a large number of voxels that share the same class as the anchor voxel, leading to a high false-negative rate. This may confuse the model training and lead to a blurred decision boundary, which could damage the segmentation performance. 
To avoid these issues, we make use of the pseudo labels computed during the forward pass to sample the negative voxels. We develop a novel \emph{confident negative sampling} strategy to sample the negative voxels. 

Given the anchor voxel $\psi_i \ (i=1,2)$, negative samples are extracted from $u^-$, which is computed from $x^-$. The pseudo label $y^-$ is calculated and the samples belonging to the same class with $\psi_i \ (i=1,2)$ are excluded. Then, we sample the negative voxels according to prediction confidence by selecting the top-K most confident samples. Such a strategy can reduce the false-negative rate during sampling and thus improves the performance compared with random sampling.

In practice, we calculate bidirectional voxel contrastive loss at every location in $u_{\phi 1}$ and $u_{\phi 2}$. Thus, the overall loss is calculated below:
\begin{equation}
\label{eqn12}
    \mathcal{L}_{bc}(u_{\phi 1}, u_{\phi 2}, u^-)=\frac{1}{N_u} \sum_{\psi_1 \in u_{\phi 1}} \mathcal{L}_{bc}(\psi_1, \psi_2),
\end{equation}
where $N_u$ denotes the number of voxels in $u_{\phi i} \ (i=1,2)$.

In summary, the whole process for training our RCPS is presented in Algorithm \ref{sec:algorithm}.
\begin{algorithm}[h]
\caption{Training Process of RCPS}
\label{sec:algorithm}

\KwIn{Segmentation network $\mathcal{F}$; Labeled dataset $\mathcal{D}_{l}$; Unlabeled dataset $\mathcal{D}_{u}$}
\KwOut{Trained segmentation network $\mathcal{F}$ for inference}
Initialize segmentation network $\mathcal{F}$\;
\While{not converged}{
Sampled batched data $x^l$ and $x^u$ from $\mathcal{D}_{l}$ and $\mathcal{D}_{u}$\;
\tcp{intensity augmentation}
Augment the data: 
$x^l_{\phi_1}, x^l_{\phi_2} \leftarrow x^l$; $x^u_{\phi_1}, x^u_{\phi_2} \leftarrow x^u$\;
\tcp{forward run}
Forward: 
$\hat{y}^{(l,u)} = \mathcal{F}(x^{(l,u)})$; $\hat{y}^{(l,u)}_{\phi_1, \phi_2} = \mathcal{F}(x^{(l,u)}_{\phi_1, \phi_2})$\;
\tcp{supervised segmentation}
Get supervised loss by Eqn. \ref{eqn1}\;
\tcp{rectified pseudo supervision}
Vanilla pseudo supervision loss $\mathcal{L}_{p}$ by Eqn. \ref{eqn4}\;
\tcp{two kinds of rectifications}
Uncertainty rectification $\mathcal{L}_{urp}$ by Eqns. \ref{eqn5} and \ref{eqn6}\;
Consistency regularization $\mathcal{L}_{cr}$ by Eqns. \ref{eqn7} and \ref{eqn8}\;
Rectified pseudo supervision loss $\mathcal{L}_{rp}$ by Eqn. \ref{eqn9}\;
\tcp{bidirectional voxel CL}
Calculate loss $\mathcal{L}_{bc}$ by Eqns. \ref{eqn10}, \ref{eqn11}, \ref{eqn12}\;
\tcp{supervised losses}
Update $\mathcal{F}$ using Eqn. \ref{eqn3}\;
\tcp{unsupervised losses}
Update $\mathcal{F}$ using Eqn. \ref{eqn2}\;
}

\end{algorithm}

\section{Experiments}
\label{sec:experiments}
In this study, two public datasets and one in-house dataset are employed to validate the proposed RCPS method, which includes the LA dataset \cite{xiong2021global}, pancreas-CT dataset \cite{nihct} and TBI dataset \cite{zhao2022tbi}. We compare the segmentation performance on the public datasets of the proposed RCPS and state-of-the-art methods, including UA-MT \cite{yu2019uncertainty}, SASSNet \cite{li2020shape}, DTC \cite{luo2021dualtask}, URPC \cite{luo2021efficient} and MC-Net$+$ \cite{wu2022mutual}, which have been discussed in Sec. \ref{sec:related_ssl}.
\subsection{Dataset and Preprocessing}
\begin{table*}[]
\centering
\caption{Performance comparison on LA dataset. The best two results are marked in bold.}
\label{table:la}
\setlength{\tabcolsep}{3mm}{
\begin{tabular}{@{}cccccc@{}}
\toprule
\multirow{2}{*}{\textbf{Method}} & \multicolumn{2}{c}{\textbf{Scans Used}} & \multicolumn{3}{c}{\textbf{Metrics}} \\ 
\cmidrule(l){2-6} 
                        & Labeled         & Unlabeled    & DSC(\%)$\uparrow$    & HD95{\scriptsize (voxel)}$\downarrow$   & ASD{\scriptsize (voxel)}$\downarrow$    \\
                        \midrule
Fully-Supervised        & 8   (10 \%)     & 0            & 79.46    & 22.18   & 6.86   \\
Fully-Supervised        & 16   (20 \%)    & 0            & 86.44    & 13.79   & 4.01   \\
Fully-Supervised        & 80   (All)      & 0            & 91.65    & 5.28    & 1.60   \\
\midrule
\midrule
UA-MT \cite{yu2019uncertainty}  & 8   (10 \%)     & 72           & 86.28    & 18.71   & 4.63   \\
SASSNet \cite{li2020shape} & 8   (10 \%)     & 72           & 85.22    & 11.18   & 2.89   \\
DTC \cite{luo2021dualtask}      & 8   (10 \%)     & 72           & 87.51    & 8.23    & 2.36   \\
URPC \cite{luo2021efficient}   & 8   (10 \%)     & 72           & 85.01    & 15.37   & 3.96   \\
MC-Net+ \cite{wu2022mutual}  & 8   (10 \%)     & 72           & \textbf{88.96} & \textbf{7.93} & \textbf{1.86}   \\
\textbf{Ours}           & 8   (10 \%)     & 72           & \textbf{90.73} & \textbf{7.91} & \textbf{2.05}   \\
\midrule
\midrule
UA-MT \cite{yu2019uncertainty}  & 16   (20 \%)    & 64           & 88.74    & 8.39    & 2.32   \\
SASSNet \cite{li2020shape} & 16   (20 \%)    & 64           & 89.16    & 8.95    & 2.26   \\
DTC \cite{luo2021dualtask}   & 16   (20 \%)    & 64           & 89.52    & 7.07    & 1.96   \\
URPC \cite{luo2021efficient}  & 16   (20 \%)    & 64           & 88.74    & 12.73   & 3.66   \\
MC-Net+ \cite{wu2022mutual}  & 16   (20 \%)    & 64           & \textbf{91.07} & \textbf{5.84} & \textbf{1.67}   \\
\textbf{Ours}           & 16   (20 \%)    & 64           & \textbf{91.21} & \textbf{6.54} & \textbf{1.81}   \\ \bottomrule
\end{tabular}}
\end{table*}

\begin{figure*}[ht]
	\centering
	\includegraphics[width=\textwidth]{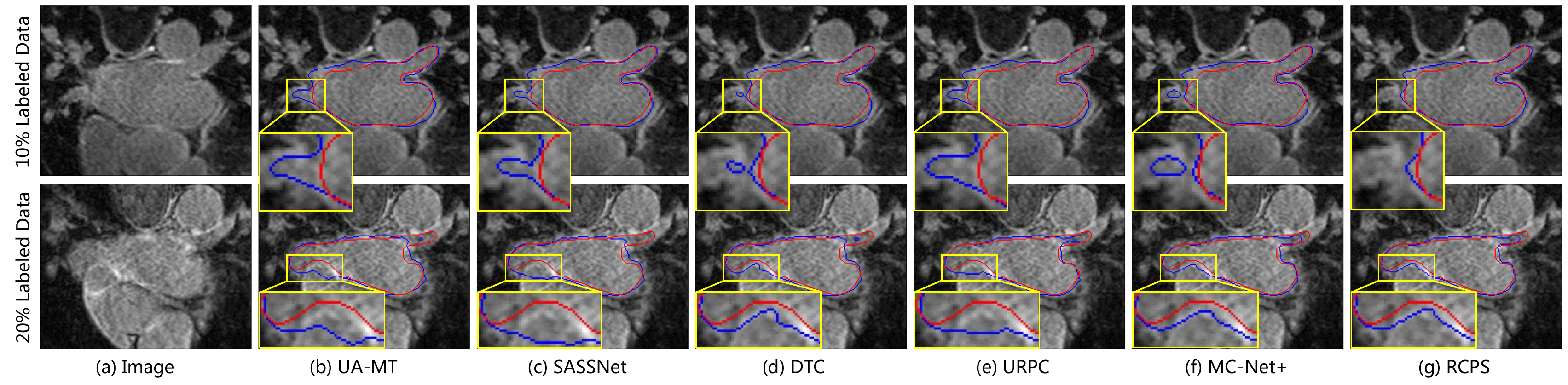} 
	\caption{Visual comparison with other methods on LA dataset. The \textcolor{red}{red} lines denote the ground truth and the \textcolor{blue}{blue} lines denote the predictions.}
	\label{fig:la}
\end{figure*}
\subsubsection{LA Dataset}
The LA dataset was the benchmark dataset for the 2018 Atrial Segmentation Challenge, which includes 100 gadolinium-enhanced labeled MR scans with an isotropic resolution of 0.625~mm × 0.625~mm × 0.625~mm. Since the annotations of the test set in LA are not available, we use the fixed data split used in previous works \cite{yu2019uncertainty, wu2022mutual, luo2022semi}, where 80 samples are used for training and the rest 20 are for validation. Then, performance comparison with other models with the same validation set is reported for fair comparison. Following existing works, two typical semi-supervised settings (\ie, training with 10\% labeled data and training with 20\% labeled data) have been conducted in our experiments.
\subsubsection{Pancreas-CT Dataset}
The pancreas-CT dataset is made public by the National Institutes of Health Clinical Center, which contains 82 3D abdominal contrast-enhanced CT scans collected with Philips and Siemens MDCT scanners, with a fixed in-plane resolution of 512 × 512 and varying intra-slice spacing from 1.5 to 2.5 mm. The data split is fixed with previous works \cite{luo2022semi}: 62 samples are used for training, and performance is reported on the rest 20 samples. For necessary preprocessing, the Hounsfield Units (HU) in all the CT scans have been rescaled, with a window level of 75 and a window width of 400. Then we resample all the scans into an isotropic resolution of 1.0 mm × 1.0 mm × 1.0 mm. We apply the same settings (\ie, training with 10\% labeled data and training with 20\% labeled data) as LA dataset in the experiments.
\subsubsection{TBI Dataset}
The TBI dataset is obtained from more than 100 patients with traumatic brain injuries at Huashan Hospital, Fudan University. Informed consent was obtained from all patients for the use of their information, medical records, and MRI data. All MR images are acquired on a 3T Siemens MR scanner and have an isotropic resolution of 1.0~mm × 1.0~mm × 1.0~mm. The dataset contains 42 labeled T1-weighted scans and 123 unlabeled T1-weighted scans. The labels include 17 consciousness-related ROIs (IR, IL, TR, TL, ICRA, ICRP, ICLA, ICLP, CRA, CRP, CLA, CLP, MCR, MCL, IPL, IPR, B) according to \cite{wu2018white}. All scans are linearly registered to MNI152 T1-weighted template using FSL \cite{smith2004advances}. Histogram equalization has been applied to ensure the stability of intensity contrast in the images. Different from the two benchmark datasets described above, we introduce the TBI dataset to explore the effectiveness of the proposed RCPS compared with previous works \cite{ren2020robust, qiao2021robust}. We perform 5-fold cross-validation on the labeled subset and report the performance on all the labeled scans, following our previous works.
\subsection{Implementation Details}
\subsubsection{Data Loading}
\begin{table*}[]
\centering
\caption{Performance comparison on the pancreas-CT dataset. The best two results are highlighted in bold.}
\label{table:pancreas}
\setlength{\tabcolsep}{3mm}{
\begin{tabular}{@{}cccccc@{}}
\toprule
\multirow{2}{*}{\textbf{Method}} & \multicolumn{2}{c}{\textbf{Scans Used}} & \multicolumn{3}{c}{\textbf{Metrics}} \\ 
\cmidrule(l){2-6} 
                        & Labeled         & Unlabeled    & DSC(\%)$\uparrow$    & HD95{\scriptsize (voxel)}$\downarrow$   & ASD{\scriptsize (voxel)}$\downarrow$    \\
                        \midrule
Fully-Supervised        & 6   (10 \%)     & 0            & 55.20   & 30.62   & 10.54   \\
Fully-Supervised        & 12   (20 \%)    & 0            & 72.38   & 20.64   & 5.41    \\
Fully-Supervised        & 62   (All)      & 0            & 83.89   & 5.08    & 2.00    \\
\midrule
\midrule
UA-MT \cite{yu2019uncertainty}   & 6   (10 \%)     & 56           & 66.44   & 17.04   & 3.03    \\
SASSNet \cite{li2020shape}  & 6   (10 \%)     & 56           & 68.97   & 18.83   & \textbf{1.96}    \\
DTC \cite{luo2021dualtask}       & 6   (10 \%)     & 56           & 66.58   & \textbf{15.46}   & 4.16    \\
URPC  \cite{luo2021efficient}    & 6   (10 \%)     & 56           & 73.53   & 22.57   & 7.85    \\
MC-Net+ \cite{wu2022mutual}   & 6   (10 \%)     & 56           & 70.00   & 16.03   & 3.87    \\
Multi-scale MC-Net+  \cite{wu2022mutual}  & 6   (10 \%)    & 60           & \textbf{74.01}   & \textbf{12.59}    & 3.34    \\
\textbf{Ours}           & 6   (10 \%)     & 56           & \textbf{76.62}   & 16.32   & \textbf{3.01}    \\
\midrule
\midrule
UA-MT  \cite{yu2019uncertainty}  & 12   (20 \%)    & 50           & 76.10   & 10.84   & 2.43    \\
SASSNet \cite{li2020shape} & 12   (20 \%)    & 50           & 76.39   & 11.06   & \textbf{1.42}    \\
DTC  \cite{luo2021dualtask}      & 12   (20 \%)    & 50           & 76.27   & 8.70    & 2.20    \\
URPC  \cite{luo2021efficient}   & 12   (20 \%)    & 50           & 80.02 & 8.51 & 1.98    \\
MC-Net+  \cite{wu2022mutual}  & 12   (20 \%)    & 50           & 79.37   & 8.52    & \textbf{1.72}    \\
Multi-scale MC-Net+  \cite{wu2022mutual}  & 12   (20 \%)    & 50           & \textbf{80.59}   & \textbf{6.47}   & 1.74    \\
\textbf{Ours}           & 12   (20 \%)    & 50           & \textbf{81.59} & \textbf{7.50} & 2.03    \\ \bottomrule
\end{tabular}}
\end{table*}

\begin{figure*}[ht]
	\centering
    \includegraphics[width=\textwidth]{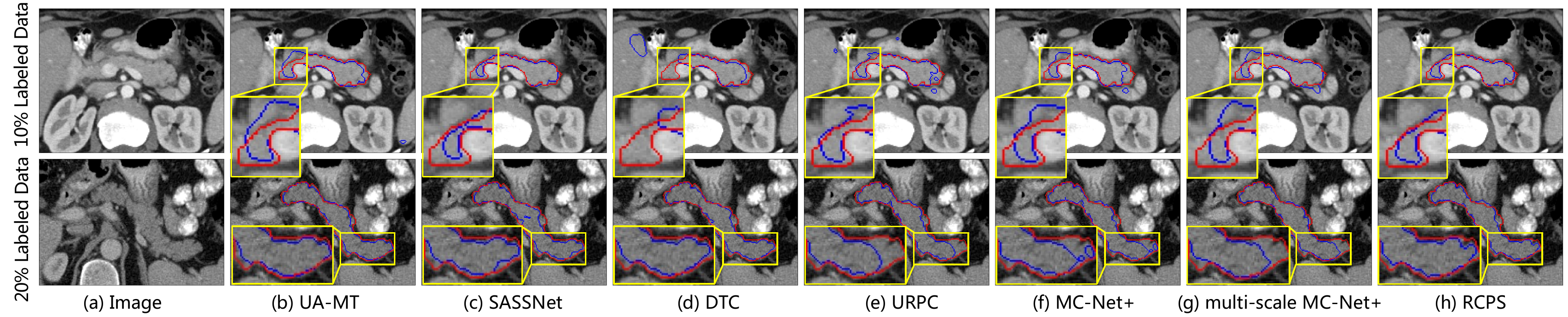} 
	\caption{Visual comparison with other methods on the pancreas-CT dataset. The \textcolor{red}{red} lines denote the ground truth and the \textcolor{blue}{blue} lines denote the predictions.}
	\label{fig:pancreas}
\end{figure*}

\begin{table}[]
\centering
\caption{Performance comparison on TBI dataset.}
\label{table:tbi}
\setlength{\tabcolsep}{3mm}{
\begin{tabular}{@{}cc@{}}
\toprule
\multirow{2}{*}{\textbf{Method}}  & \textbf{Metrics}\\ \cmidrule(l){2-2} 
                         & DSC(\%)$\uparrow$   \\
                         \midrule
U-Net \cite{ronneberger2015u}        & 61.64   ± 21.60 \\
nnU-Net \cite{isensee2021nnu}        & 63.37   ± 23.62 \\
Ren   et al. \cite{ren2020robust} & 67.19   ± 17.18 \\
Qiao et al. \cite{qiao2021robust} & 69.03   ± 14.85 \\
\textbf{Ours}            & \textbf{71.88 ± 15.90} \\ \bottomrule
\end{tabular}}
\end{table}
All 3D scans are first normalized into zero mean and unit variance before feeding into networks. Following \cite{yu2019uncertainty, luo2021dualtask, wu2022mutual}, for LA and pancreas-CT datasets, all scans are first center-cropped around the ROI, with enlarged margins of 25 voxels in all the spatial axes. Since the training of 3D data is computationally demanding, we crop all the training data into patches during training. The training patch size is set to 112 × 112 × 80 for LA, and 96 × 96 × 96 for pancreas-CT and TBI, according to the previous works in \cite{yu2019uncertainty, luo2021dualtask, wu2022mutual}. For data augmentation, we only adopt random grid distortion which involves perturbing the pixel grid of an input image with random deformations to generate additional training samples. By applying random grid distortion, the augmented training samples exhibit various deformations while its authenticity is also guaranteed. During inference, we adopt a patch-based pipeline with a sliding window strategy to merge the patch predictions and acquire the final outputs.
We utilize MONAI library \cite{monai} to accelerate our data pipeline, including data loading, data augmentation, and patch-based inference.

\begin{figure*}[ht]
	\centering
	\includegraphics[width=\textwidth]{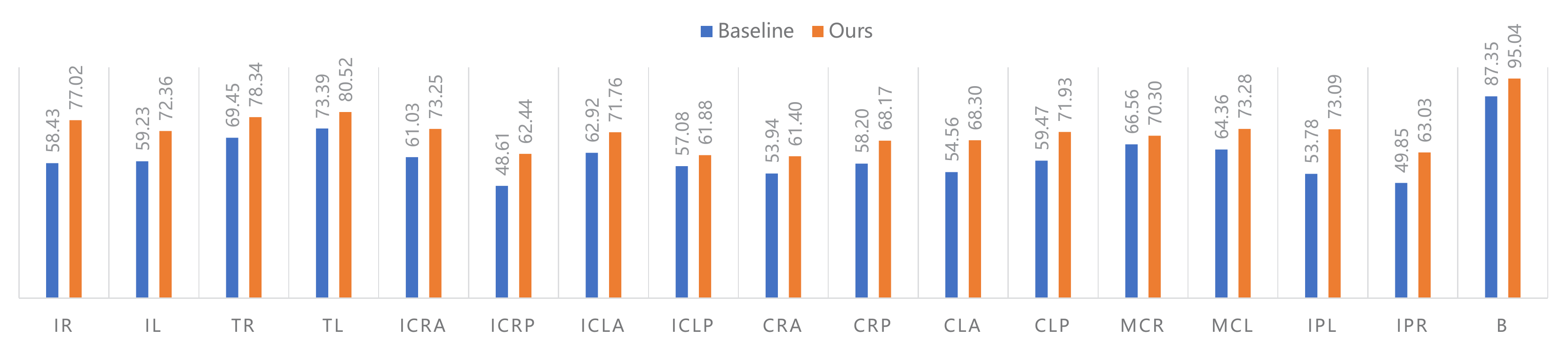} 
	\caption{Performance comparison of different brain regions with U-Net baseline on TBI dataset.}
	\label{fig:tbi_performance}
\end{figure*}
\begin{figure*}[ht]
	\centering
	\includegraphics[width=\textwidth]{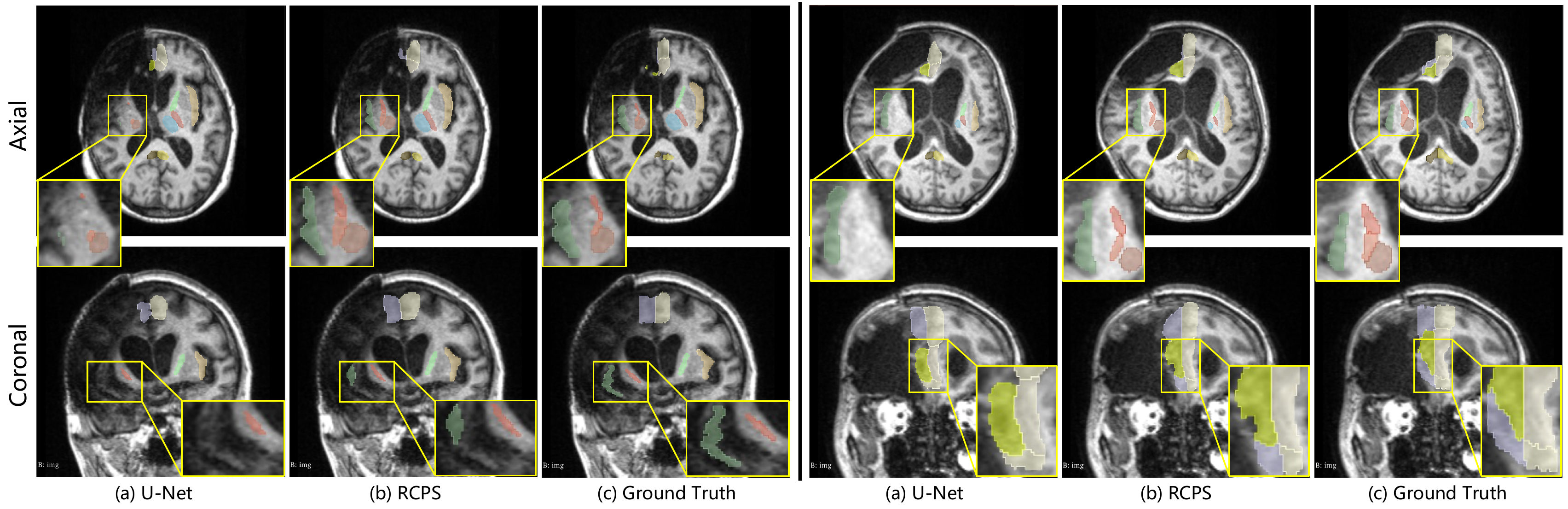} 
	\caption{Visual comparison with U-Net on TBI dataset.}
	\label{fig:tbi}
\end{figure*}

\subsubsection{Training Configurations}
The proposed RCPS is implemented with PyTorch 1.12.1 framework on a Debian server. Distributed data parallel computation with 2 NVIDIA TITAN RTX GPUs is used to accelerate the training process. PyTorch native automatic mixed-precision training is employed to save video memory usage. Gradient checkpointing is also introduced for the computation of bidirectional voxel contrastive loss to avoid OOM error. The number of negative sampling is set to 400. The network backbone is a 3D U-Net. The total batch size is set to 4, where each mini-batch contains 2 labeled images and 2 unlabeled images. All of the segmentation tasks are optimized by an SGD optimizer with a momentum of 0.9 and weight decay of $1 \times 10^{-4}$. The initial learning rate is set to $1 \times 10^{-2}$ and is slowly reduced with polynomial strategy. We have trained the network for 200 epochs for LA and 400 epochs for pancreas-CT and TBI. For hyperparameter settings, we set $T=0.5$ and $\tau=0.1$ throughout the experiments, following \cite{zhong2021pixel} and \cite{wang2021exploring}. $\alpha$ and $\beta$ are task-specific hyperparameters. Following \cite{zhong2021pixel} and \cite{you2022simcvd}, we set $\beta=0.1$ throughout the experiments. Meanwhile, we empirically set $\alpha=0.1$ for LA dataset and $\alpha=0.2$ for pancreas-CT and TBI datasets according to the difficulty levels of those segmentation tasks. Note that $\alpha$ and $\beta$ are set as a time-dependent Gaussian warming-up function to stabilize the training, in line with previous benchmarks \cite{luo2022semi, wu2022mutual}.
Following previous works on these benchmarks \cite{yu2019uncertainty, li2020shape, luo2021dualtask, luo2021efficient, luo2022semi, wu2022mutual}, we use Dice similarity coefficients (DSC), 95\% Hausdorff distance (HD95), and average surface distance (ASD) to evaluate the segmentation methods on LA dataset and pancreas-CT dataset. For TBI dataset, we report the mean value and standard deviation of Dice similarity coefficients on the labeled scans during cross-validation, following our previous works \cite{ren2020robust, qiao2021robust}.
\subsection{Results}
\subsubsection{LA Dataset}
The quantitative results of LA dataset are presented in Table \ref{table:la}. Fully-supervised 3D U-Net is used as the baseline compared with semi-supervised learning. Overall, in both the 10\% setting and 20\% setting, the proposed RCPS yields the best segmentation DSC and the best or second best HD95 or ASD performance. The proposed RCPS improves the segmentation DSC from 79\% to more than 90\% with only 10\% labeled data, yielding a significant performance leap compared with the fully-supervised method. When the labeled data is increased to 20\%, the segmentation performance is improved to 91.21\%, close to the upper bound of fully-supervised learning (91.65\%). In both settings, the proposed RCPS outperforms other cutting-edge semi-supervised methods, which demonstrates the superiority of our method. In Fig. \ref{fig:la}, we have visualized the segmentation results on LA dataset in both two settings. Compared with other methods, the proposed RCPS generates a complete segmentation prediction and provides more fine details in small structures. 
\subsubsection{Pancreas-CT Dataset}
The quantitative results of the pancreas-CT dataset are shown in Table \ref{table:pancreas}. Fully-supervised methods with 10\%, 20\%, and all labeled data are introduced as benchmarks for evaluating the effectiveness of semi-supervised learning. Compared with LA dataset, the pancreas-CT segmentation is a more difficult task, which requires better design in semi-supervised algorithms to yield satisfactory results. Despite the challenges in the segmentation task, the proposed RCPS still yields promising performance in both of the two settings, achieving best segmentation DSC in both 10\% and 20\% settings, as well as promising results in HD95 and ASD.
When training with only 10\% labeled data, the proposed RCPS improves the segmentation performance by a large margin, raising from 55\% to 76\%, and outperforms all other semi-supervised methods. When the amount of labeled data is increased to 20\%, we manage to improve the segmentation performance to 81.59\%, which is approaching the upper bound of fully-supervised learning (83.89\%) as well. 
Also, it should be noted that RCPS, as a model built upon data-based perturbations, does not require any further computational cost during inference. Thus, we achieve a more sophisticated balance between performance and efficiency compared with the multi-scale method in \cite{wu2022mutual}, despite slightly less favorable performance in distance-related metrics.
Similar to the results in LA dataset, the segmentation visualization reveals that the proposed RCPS tends to avoid predicting small isolated components during segmentation, and generates more reasonable segmentation maps for tiny structures, which is shown in Fig. \ref{fig:pancreas}.
\subsubsection{TBI Dataset}
Apart from binary segmentation in LA and pancreas-CT datasets, the proposed RCPS has been evaluated on the in-house TBI dataset. Quantitative results in Table \ref{table:tbi} show that the proposed RCPS is still competitive in multi-class segmentation tasks. With the aid of unlabeled data, the segmentation performance of a common U-Net has improved by more than 16\% (61.64\% to 71.88\% in DSC), outperforming all previous works \cite{ren2020robust} \cite{qiao2021robust}. Semi-supervised methods significantly improve segmentation performance in all brain regions, especially those heavily affected by the injuries, which is shown in Fig. \ref{fig:tbi_performance}. Furthermore, we visualize several segmentation results in Fig. \ref{fig:tbi}. U-Net trained under semi-supervised settings is capable to segment the structures usually missed by the fully-supervised U-Net, and produces segmentation maps of a much higher quality.
\begin{table}[]
\caption{Ablation studies on different losses. The ratio of labeled data is set to 10 \%.}
\label{table:ablation_loss}
\centering
\scriptsize
\setlength{\tabcolsep}{0.7mm}{
\begin{tabular}{@{}cccccccc@{}}
\toprule
\multirow{2}{*}{$\mathcal{L}_{rp}$} & \multirow{2}{*}{$\mathcal{L}_{bc}$} & \multicolumn{3}{c}{LA} & \multicolumn{3}{c}{Pancreas-CT} \\ \cmidrule(l){3-8} 
                     &                      & DSC(\%)$\uparrow$ & HD95{\tiny (voxel)}$\downarrow$ & ASD{\tiny (voxel)}$\downarrow$ & DSC(\%)$\uparrow$ & HD95{\tiny (voxel)}$\downarrow$ & ASD{\tiny (voxel)}$\downarrow$  \\ \midrule
                     &                      & 78.46  & 23.18  & 6.86 & 55.20    & 30.62   & 10.54   \\
\checkmark           &                      & 86.21  & 17.35  & 3.70 & 73.14    &\textbf{13.72} & 3.10 \\
                     & \checkmark           & 86.53  & 18.28  & 3.98 & 62.49    & 33.00   & 6.31    \\
\checkmark           & \checkmark           & \textbf{90.73} & \textbf{7.91} & \textbf{2.05} & \textbf{76.62} & 16.32 & \textbf{3.01}    \\ \bottomrule
\end{tabular}}
\end{table}
\begin{table}[]
\centering
\scriptsize
\caption{Ablation studies on rectification in pseudo supervision. \emph{UE} denotes uncertainty estimation and \emph{CR} denotes consistency regularization. The ratio of labeled data is set to 10 \%.}
\label{table:ablation_rectification}
\setlength{\tabcolsep}{0.7mm}{
\begin{tabular}{@{}ccccccc@{}}
\toprule
\multirow{2}{*}{LA}          & \multicolumn{3}{c}{w/o   UE} & \multicolumn{3}{c}{w/   UE} \\ \cmidrule(l){2-7} 
                             & DSC(\%)$\uparrow$ & HD95{\tiny (voxel)}$\downarrow$ & ASD{\tiny (voxel)}$\downarrow$ & DSC(\%)$\uparrow$ & HD95{\tiny (voxel)}$\downarrow$ & ASD{\tiny (voxel)}$\downarrow$       \\ \midrule
w/o   CR            & 89.42       & 13.46       & 2.90      & 90.12       & 10.34      & 2.37      \\
w/   CR             & 89.94       & 8.60        & 2.11      & \textbf{90.73} & \textbf{7.91} & \textbf{2.05}      \\
\midrule
\midrule
\multirow{2}{*}{Pancreas-CT} & \multicolumn{3}{c}{w/o   UE} & \multicolumn{3}{c}{w/   UE} \\ \cmidrule(l){2-7} 
                             & DSC(\%)$\uparrow$ & HD95{\tiny (voxel)}$\downarrow$ & ASD{\tiny (voxel)}$\downarrow$ & DSC(\%)$\uparrow$ & HD95{\tiny (voxel)}$\downarrow$ & ASD{\tiny (voxel)}$\downarrow$       \\ \midrule
w/o   CR            & 72.18       & 23.13       & 5.18      & 75.49       & \textbf{14.27} & 3.09      \\
w/   CR             & 74.25       & 17.91       & 3.72      & \textbf{76.62} & 16.32      & \textbf{3.01}      \\ \bottomrule
\end{tabular}}
\end{table}
\subsection{Ablation Studies}
In this section, we validate the effectiveness of the rectified pseudo supervision and bidirectional contrastive learning in the proposed RCPS method. The experiments are carried out on the LA dataset and the pancreas-CT dataset, with the labeled data ratio being 10\%.
\subsubsection{The Effects of $\mathcal{L}_{rp}$ and $\mathcal{L}_{bc}$}
The proposed RCPS does not introduce any extra modification to the U-Net architecture. The semi-supervised segmentation is conducted by introducing two important losses during training. We report the effectiveness of the two loss functions in Table \ref{table:ablation_loss}. Experimental results indicate that both of the losses contribute to the performance gain in semi-supervised learning, and the maximum performance gain is obtained when the losses are combined together.
\subsubsection{The Effects of Rectification in $\mathcal{L}_{rp}$}
Two kinds of rectification have been introduced in pseudo supervision to ensure stable training and consequent promising segmentation performance. First, the uncertainty estimation between the pseudo label and the predictions of augmented views, and second the consistency regularization between the two different augmented views. The quantitative results in Table \ref{table:ablation_rectification} show that both the rectifications could improve the quality of pseudo supervision. Uncertainty estimation contributes more to performance elevation in pseudo supervision, compared to consistency regularization. However, a combined rectification of both operations leads to optimal results.
\begin{table}[]
\centering
\scriptsize
\caption{Ablation studies on the number of negative samples $N$ in bidirectional voxel contrastive loss. The ratio of labeled data is set to 10 \%.}
\label{table:ablation_sample}
\setlength{\tabcolsep}{1mm}{
\begin{tabular}{@{}ccccccc@{}}
\toprule
\multirow{2}{*}{$N$} & \multicolumn{3}{c}{LA} & \multicolumn{3}{c}{Pancreas-CT} \\ \cmidrule(l){2-7} 
                                & DSC(\%)$\uparrow$ & HD95{\tiny (voxel)}$\downarrow$ & ASD{\tiny (voxel)}$\downarrow$ & DSC(\%)$\uparrow$ & HD95{\tiny (voxel)}$\downarrow$ & ASD{\tiny (voxel)}$\downarrow$     \\ \midrule
0                                & 86.21  & 17.35  & 3.70 & 73.14     & 13.72     & 3.10    \\
100                              & 90.59  & 8.04   & 2.18 & 75.42     & 16.51     & 3.23    \\
200                              & 90.52  & 7.98   & 2.23 & 76.48     & 17.78     & 3.89    \\
300                              & 90.70  & \textbf{7.49} & 2.09 & 76.59     & 16.68     & 3.17    \\
400                              & \textbf{90.73}  & 7.91   & \textbf{2.05} & \textbf{76.62}  & \textbf{16.32}  & \textbf{3.01}    \\ \bottomrule
\end{tabular}}
\end{table}
\begin{table}[]
\centering
\scriptsize
\caption{Ablation studies on sampling strategy in bidirectional voxel contrastive loss. \emph{Rand.} denotes random sampling and \emph{Conf. Neg.} denotes confident negative sampling. The ratio of labeled data is set to 10 \% and the number of negative samples is set to 400.}
\label{table:ablation_strategy}
\setlength{\tabcolsep}{0.7mm}{
\begin{tabular}{@{}ccccccc@{}}
\toprule
\multirow{2}{*}{Strategy} & \multicolumn{3}{c}{LA} & \multicolumn{3}{c}{Pancreas-CT} \\ \cmidrule(l){2-7} 
                                   & DSC(\%)$\uparrow$ & HD95{\tiny (voxel)}$\downarrow$ & ASD{\tiny (voxel)}$\downarrow$ & DSC(\%)$\uparrow$ & HD95{\tiny (voxel)}$\downarrow$ & ASD{\tiny (voxel)}$\downarrow$     \\ \midrule
Rand.                  & 86.59  & 14.91  & 3.21 & 73.48     & 23.06     & 3.95    \\
Conf. Neg.     & \textbf{90.73}  & \textbf{7.91} & \textbf{2.05} & \textbf{76.62}  & \textbf{16.32}  & \textbf{3.01}    \\ \bottomrule
\end{tabular}}
\end{table}
\subsubsection{The Effects of Confident Negative Sampling}
We discuss the effectiveness of confident negative sampling by conducting two series of experiments. We first investigate the effects of the number of negative samples $N$ in contrastive learning in Table \ref{table:ablation_sample}. Experimental results show that a larger bank of negative samples leads to greater performance gain, but the performance boost is marginal when $N$ is greater than 100. Although we set $N$ to 400 throughout the experiments, setting $N$ from 100 to 200 can still yield good performance. This is significant in practice as the computation of contrastive loss is rather demanding during training. When we set $N=400$, the mixed precision training process can take up more than 20 GB VRAM, which can set a barrier to its training on some older GPUs. When $N=100$, VRAM usage drops to approximately 10 GB with little performance compromise, making the proposed approach accessible to most modern GPUs.

We also explore the effects of the proposed sampling strategy compared with completely random sampling. The results in Table \ref{table:ablation_strategy} indicate that random sampling does not produce performance gains despite the introduction of bidirectional contrastive learning of voxels, demonstrating the effectiveness of the proposed sampling strategy.

\section{Discussion and Conclusion}
\label{sec:conclusion}
In this work, we proposed a novel RCPS (rectified contrastive pseudo supervision) method for semi-supervised segmentation in medical images. We introduced two kinds of rectifications to improve the effectiveness of pseudo supervision, which are the uncertainty estimation between the pseudo label and the predictions of augmented views, and the consistency regularization between the predictions from the two augmented views. Furthermore, we implemented a bidirectional voxel contrastive loss in the feature space to improve the class separability directly in the segmentation. The introduction of bidirectional voxel contrastive loss ensures both a large inter-class distance and a tight intra-class clustering in the feature space, leading to further performance gain. Experimental results revealed that the proposed RCPS improves the segmentation performance by a large margin compared with the fully-supervised model trained with few annotated samples. Furthermore, our method has outperformed state-of-the-art methods in public benchmarks.

Despite the success of the proposed RCPS method in semi-supervised medical image segmentation, there are potential limitations that should be addressed in future works: First, the constructed model of RCPS is mainly based on intensity-based perturbations of the inputs. Such data-based perturbations rely on the manual design of the augmentation pipelines, such as the selection of augmentations and the strength of perturbations. Furthermore, we have found in experiments that despite the satisfactory performance of RCPS in terms of DSC scores, distance-related metrics like HD95 or ASD are not as promising as DSC. The plausible reason is that no related loss functions are incorporated into our method, which limits the capacity of our model to also achieve better performance in these metrics.

Based on the aforementioned limitations in our study, the corresponding remedies and our future extensions are listed as follows: 1) We will explore parameterized augmentations upon the input data to resolve the issues during the manual design of data-based perturbations; 2) We will further introduce model-based perturbations and distance-related loss functions in our framework, which could further improve the segmentation performance especially in terms of distance-related metrics; 3) We will further apply RCPS method to other semi-supervised segmentation tasks in the medical image domain, and will extend it to 2D medical image segmentation as well, to demonstrate the effectiveness of proposed RCPS as a universal semi-supervised segmentation method.

\bibliography{references}
\end{document}